\documentclass[letterpaper, 10 pt, conference]{ieeeconf}  

\IEEEoverridecommandlockouts                              

\usepackage{todonotes}
\usepackage{bm}
\usepackage{amssymb}
\usepackage{amsmath}
 \usepackage{float}
\newcommand*{\vect}[1]{\bm{#1}}
\newcommand*{\vectd}[1]{\dot{\bm{#1}}}
\newcommand*{\vectdd}[1]{\ddot{\bm{#1}}}

\usepackage{mathptmx}
\usepackage{times}
\usepackage[table]{xcolor}
\usepackage{siunitx}
\usepackage{fancyhdr}
\usepackage{ifthen}

\renewcommand{\baselinestretch}{0.933}
\title{\LARGE \bf
Control of  Fully Actuated Aerial Vehicles: A Comparison of Model-based and Sensor-based Dynamic Inversion
}

\author{Ali Sidar Yilmaz$^{1}$, Buday Turan$^{1}$, Lukas Pries$^{1}$, Markus Ryll$^{1}$
\thanks{$^{1}$All authors are with the Professorship of Autonomous Aerial Systems, School of Engineering and Design, Technical University of Munich, Lise-Meitner-Str.\ 9, 85521 Ottobrunn, Germany ({\tt\small sidar.yilmaz@tum.de, buday.turan@tum.de, lukas.pries@tum.de, markus.ryll@tum.de}).}%
\thanks{Experiment video: {\tt \footnotesize www.youtube.com/watch?v=n7Zoibjk73w}}%
}

\begin{document}
\maketitle

\newboolean{preprint}
\setboolean{preprint}{true}

\ifthenelse{\boolean{preprint}}
{%
    \setlength{\headheight}{18pt}
    \renewcommand{\headrulewidth}{0pt}
    \pagestyle{fancy}
    \fancyhf{}

    \fancyhead[L]{%
      \vspace{1.8em}%
      \small INTERNATIONAL CONFERENCE ON UNMANNED AIRCRAFT SYSTEMS 2026. PREPRINT VERSION. ACCEPTED MAY 2026%
    }

    \thispagestyle{fancy}
}
{%
    \thispagestyle{empty}
    \pagestyle{empty}
}

\begin{abstract}

Fully actuated multirotor platforms decouple translational force generation from vehicle attitude, enabling independent control of position and orientation and shifting performance limitations from attitude authority to actuator dynamics and control effectiveness. This paper compares a model-based nonlinear dynamic inversion controller (geometric NDI) with a sensor-based incremental dynamic inversion controller (INDI) on a fixed-tilt fully actuated hexarotor. Both controllers share an identical outer-loop structure and are both executed at 500~Hz; therefore, performance differences can be attributed primarily to the inversion strategy. Controller performance is evaluated in five experiments covering attitude step tracking under nominal conditions and under a 50\% mismatch in the rotor force coefficient, hover disturbance rejection under an external lateral load, waypoint tracking in the presence of wind gust disturbances, reduced control frequency, and injected sensor degradation. The results show that INDI offers clear advantages under parameter mismatch, gust disturbances, and sensor degradation, and maintains lower position errors across the controller-frequency sweep. However, its advantages are not universal: geometric NDI yields better attitude tracking at reduced control frequencies. To the authors’ best knowledge, this work presents the first experimental validation of a full pose tracking INDI controller with decoupled translational and rotational dynamics. These findings highlight the trade-off between measurement-based and model-based inversion for robust control and rapid deployment of fully actuated UAVs.
\end{abstract}
\section{INTRODUCTION}
Incremental Nonlinear Dynamic Inversion (INDI) and model-based Nonlinear Dynamic Inversion (NDI) are widely used control approaches for multirotor platforms. In both cases, tracking errors can be defined directly on geometric manifolds such as 
$SO(3)$ and $SE(3)$, enabling coordinate-free representations of attitude and pose with well-established stability properties \cite{Lee2010}. Although the error definitions used for the INDI controller are not formulated directly on $ SO (3) $, the main difference between the two methods does not lie in the error formulation but in how the desired accelerations are translated into control wrenches or actuator commands. Model-based NDI relies on inverting a dynamics model to compute the required input, whereas INDI uses incremental dynamics and high-rate sensor measurements to update the motor input. This incremental formulation reduces dependence on an accurate model and provides strong disturbance rejection \cite{Smeur2018,Grondman2018}. Both approaches have been successfully demonstrated on conventional multirotors, which are inherently underactuated \cite{Lee2010,Goodarzi2015,Zhang2023}. 

For underactuated multirotors, translational force generation is coupled to the attitude dynamics. Consequently, the performance of both INDI and geometric NDI is fundamentally limited by the attitude control authority and rotational dynamics rather than by actuator force capabilities. Fully actuated multirotor platforms remove this coupling by enabling direct control of the full body wrench \cite{Ryll2014,Kamel2018}. This allows position and attitude to be controlled independently. In this case, the controller performance is no longer constrained by attitude dynamics but instead by actuator-related limitations such as motor dynamics and rotor thrust coefficients \cite{Seshasayanan2026}. 

To the authors’ best knowledge, this work presents the first experimental validation of a full pose tracking INDI controller on a fully actuated multirotor platform with decoupled translational and rotational dynamics. In addition, the paper provides a systematic and experimentally controlled comparison between model-based geometric NDI and sensor-based INDI on the same fixed-tilt fully actuated hexarotor. By employing an identical outer-loop structure and identical execution rates, performance differences can be attributed directly to the respective inversion strategies rather than to implementation details. 

\section{Related Work}
\subsection{Drone Configurations}
Multirotor UAVs can be grouped based on their actuation capabilities. The most common distinction is between underactuated and fully actuated platforms. This distinction directly affects how forces and moments can be generated.

Conventional quadrotors and hexarotors with fixed, vertically oriented thrust vectors are underactuated systems. Thrust is produced only along a single body-fixed axis. As a result, horizontal motion requires tilting the vehicle, which couples position and attitude dynamics. This coupling limits performance, especially during aggressive maneuvers or in the presence of disturbances. Such configurations have been widely studied and form the standard model for multirotor control research \cite{Mellinger2011,Faessler2018}.

To overcome this limitation, several fully actuated multirotor configurations have been proposed. One approach is to arrange rotors with non-parallel thrust directions, allowing independent generation of forces and moments. Early omnidirectional designs demonstrated full wrench controllability without relying on attitude changes, enabling decoupled position and orientation control \cite{Ryll2017,Kamel2018}. Other platforms achieve full actuation by increasing the number of rotors or by carefully selecting rotor orientations while keeping the mechanical design rigid \cite{Brescianini2018}.

Overall, fully actuated fixed-tilt multirotors offer a good balance between control authority and mechanical simplicity. This makes them a suitable choice for evaluating advanced control strategies under reduced dynamic coupling.

\subsection{Controller}
INDI was developed to address a key limitation of model-based NDI: its reliance on accurate system models. By leveraging high-rate sensor measurements, INDI reduces dependence on model knowledge, resulting in improved robustness to modeling errors \cite{SieberlingINDI}. Moreover, the reduced model dependence allows rapid deployment, making it well-suited for experimental platforms.

INDI has been implemented in various flight control applications. Grondman et al. \cite{Grondman2018} implemented INDI on a CS-25-certified aircraft for the first time. They compared INDI's performance with a classical NDI controller and found that INDI was more robust to model uncertainties. Simplício et al. \cite{INDIheli1}  present an application of INDI on a single main rotor and tail rotor helicopter, and the robustness of the controller was successfully tested against aerodynamic uncertainties and a tail rotor malfunction case in simulations. Smeur et al. \cite{INDImulticopter2} demonstrate an adaptive INDI scheme that estimates control effectiveness online for micro aerial vehicles. Their tests demonstrated that complex aerodynamic models are not required, and the controller adapts to new configurations with ease. Acquatella et al. \cite{INDIspace1} presented attitude tracking and disturbance rejection using INDI for spacecraft applications. The INDI controller was tested against NDI with PI control, and INDI demonstrated improved performance under the influence of external disturbances, time delays, and parametric uncertainty. Hachem et al. \cite{Hachem2025} demonstrated a cascaded full-pose tracking INDI/$\mathcal{H}\infty$ control structure for over-actuated multirotors. The outer INDI/$\mathcal{H}\infty$ guidance loop ensures reference position and attitude tracking by generating feasible incremental force and attitude commands, while the inner INDI controller generates incremental torque commands based on the attitude commands from the outer loop. The implementation was evaluated in simulations, and the results demonstrated that the proposed controller architecture achieves successful tracking. Blaha et al. \cite{Blaha2024} presented an INDI attitude controller with an NDI position outer loop for quadrotors. Through simulations and experiments, their control and identification algorithms were successfully validated for throw recovery and position control. Many additional INDI implementations with simulation and flight-test results are reported in the literature \cite{INDIVAACSmith,INDImulticopter1,INDIspace2}. An overview of INDI and its variants is provided in \cite{INDIoverview}. 

Geometric control can be regarded as the manifold-consistent formulation of model-based nonlinear dynamic inversion (NDI), where tracking errors are defined directly on $SO(3)$ and $SE(3)$ and the rigid-body dynamics are inverted to obtain the required control wrench. As in classical NDI, the controller relies on a dynamics model to cancel nonlinearities and impose desired closed-loop error dynamics. The main distinction lies in the representation: instead of expressing attitude in local coordinates, geometric control defines configuration errors intrinsically on the rotation and pose manifolds, yielding coordinate-free tracking laws with almost-global stability properties that remain valid for large rotations and aggressive maneuvers \cite{Mayhew2011,Bullo2005}.

Geometric controllers based on NDI have been extensively demonstrated on multirotor platforms. Lee et al. \cite{Lee2010} validated almost global $SE(3)$ tracking on a quadrotor in aggressive flight experiments. Mellinger et al. \cite{Mellinger2011} combined geometric tracking with minimum-snap trajectory generations for high-precision maneuvers.  Goodarzi et al. \cite{Goodarzi2013} incorporated aerodynamic drag and actuated dynamics with the same geometric inversion framework. Extensions to aerial manipulation and cooperative transport have also been developed using a geometric control approach \cite{Lee2015,Sreenath2013}. For fully actuated multirotor platforms, geometric control on wrench enables direct six-degree-of-freedom tracking and aerial interaction \cite{Ryll2014, Rajappa2015}. Foundational background on geometric rigid-body control on Lie groups is provided in \cite{Bullo2005}.

\section{Methodology}
\subsection{Geometric Control with Model Based Inversion}
The control problem addressed in this section is tracking an offline-calculated desired position and orientation trajectory. This formulation is not novel to this work, but is restated here for clarity and completeness of the paper.
\[
(\vect{p}_d(t), \vect{q}_d(t)) \in \mathbb{R}^3 \times \mathbb{S}^3,
\]
with the main body's position $\vect{p}_b \in \mathbb{R}^3$ and orientation expressed as a unit quaternion $\vect{q}_b \in \mathbb{S}^3$. However, this control approach requires a system model.

The aerodynamic forces and torques of a single propeller are commonly modeled as
\begin{align}
\vect{f}_i^{P_i} &= c_f w_i|w_i| \vect{e}_3, \quad &i=1,\dots,6,\label{eq:rotor_force}\\
\vect{\tau}_i^{P_i} &= (-1)^{i-1} c_{\tau} w_i|w_i| \vect{e}_3, \quad &i=1,\dots,6,
\end{align}
where $\vect{f}_i^{P_i}, \vect{\tau}_i^{P_i}$ are expressed in the local $i$-th propeller frame $P_i$, $c_f$ and $c_{\tau}$ are propeller geometry-dependent parameters, $w_i$ is the spinning velocity of propeller $i$, and $\vect{e}_3 = [0\ 0\ 1]^\top$.

The total aerodynamic lift force $\vect{f}^w$ in world frame and torque $\vect{\tau}^B$ in body frame are then given by
\begin{align}
\vect{f}^w(\omega_{1..6}) &= \vect{R}_B^w \sum_{i=1}^6 \vect{R}_{P_i}^B \vect{f}_i^{P_i}
= \vect{R}_B^w \vect{F}_1 \vect{u},\\
\vect{\tau}^B(\omega_{1..6}) &= \sum_{i=1}^6 \Big( \vect{p}_{B,P_i}^B \times (\vect{R}_{P_i}^B \vect{f}_i^{P_i}) + \vect{R}_{P_i}^B \vect{\tau}_i^{P_i} \Big) \notag\\
&= \vect{F}_2\vect{u},
\end{align}
where $\vect{R}_B^w$ and $\vect{R}_{P_i}^B$ are rotation matrices mapping from body to world and from propeller to body, respectively. The vector $\vect{p}_{B,P_i}^B$ denotes the position of the $i$-th propeller relative to the center of mass in body frame. Finally, the input vector is
\begin{equation}
\vect{u} = [\, w_1|w_1|, \dots, w_6|w_6| \,]^\top.
\end{equation}

The complete dynamics can now be written as
\begin{equation}
\begin{bmatrix}
m \vectdd{p} \\
\vect{J} \vectd{\omega}
\end{bmatrix}
=
\begin{bmatrix}
- m g \vect{e}_3 \\
\vect{\omega} \times \vect{J} \vect{\omega}
\end{bmatrix}
+
\begin{bmatrix}
\vect{F}_1 \\
\vect{F}_2
\end{bmatrix} \vect{u}
= \vect{f} + \vect{F} \vect{u},
\label{eq:model}
\end{equation}

where $m$ and $\vect{J}$ are the mass and inertia of the system, and $g$ is the gravitational acceleration. The translational dynamics are expressed in the inertial frame, while the rotational dynamics are expressed in the body frame.

The position and velocity errors are defined as
\begin{align}
\vect{e}_p &= \vect{p}_d - \vect{p}_b,\\
\vect{e}_v &= \vect{v}_d - \vect{v}_b,
\label{eq:pos_error}
\end{align}
where $\vect{v}_d = \vectd{p}_d$ and $\vect{v}_b = \vectd{p}_b$.  

The attitude tracking error is expressed as
\begin{equation}
\vect{e}_q = \vect{q}_d \otimes \vect{q}_b^{-1},
\label{eq:att_error}
\end{equation}
and the angular velocity tracking error as
\begin{equation}
\vect{e}_\omega = \vect{\omega}_b - \vect{R}(\vect{q}_b)^\top \vect{R}(\vect{q}_d) \vect{\omega}_d,
\label{eq:qvel_error}
\end{equation}
where $\vect{R}(\vect{q})$ is the rotation matrix corresponding to quaternion $\vect{q}$, and $\vect{\omega}_d, \vect{\omega}_b \in \mathbb{R}^3$ are the desired and current angular velocities.

Using the error terms \eqref{eq:pos_error}–\eqref{eq:qvel_error}, we define pseudo control inputs as
\begin{align}
\vect{v}_p &= \vect{k}_p \vect{e}_p + \vect{k}_v \vect{e}_v + \vectdd{p}_d, \\
\vect{v}_{att} &= \vect{k}_q \vect{e}_q + \vect{k}_\omega \vect{e}_\omega + \vectd{\omega}_d,
\end{align}
where $\vect{k}_{(\cdot)}$ are positive diagonal gain matrices.

To apply exact feedback linearization, we rewrite \eqref{eq:model} as
\begin{equation}
\vect{u} = \vect{F}^{-1}\!\left(
- \vect{f} +
\begin{bmatrix}
m I_3 & 0 \\
0 & \vect{J}
\end{bmatrix}
\begin{bmatrix}
\vect{v}_p \\
\vect{v}_{att}
\end{bmatrix}
\right),
\label{eq:NDIcontrol_law}
\end{equation}
which ensures asymptotic convergence of both the position and orientation error terms.

\subsection{Sensor-Based Inversion Controller (INDI)}
INDI derivations for SISO and MIMO systems with affine and non-affine inputs have been reported in the literature, for instance by Steinert et al. \cite{INDIoverview}. In this work, an INDI derivation for a fully actuated system is presented. Consider the system output and the desired pseudo-control input as,

\begin{equation}
    \vect{y} = \begin{bmatrix}
\vectd{p} \\
\vect{\omega}
\end{bmatrix}
\end{equation}
\begin{equation}
    \vect{\nu}_{d} = \vectd{y}_{d}.
\end{equation}

The NDI control law in \eqref{eq:NDIcontrol_law} is highly model dependent due to the $\bm{f}$ and $\bm{F}^{-1}$ terms. To reduce the model dependence, $\bm{f}$ can be replaced with the current measurements denoted by subscript '$0$' such that,
\begin{equation}
        \begin{bmatrix}
m I_3 & 0 \\
0 & \vect{J}
\end{bmatrix}\dot{\bm{y}}_0 = 
\begin{bmatrix}
m \vectdd{p}_0 \\
\vect{J} \vectd{\omega}_0
\end{bmatrix}
= \vect{f}_0 + \vect{F} \vect{u}_0
\end{equation}
       
\begin{equation}
     \bm{f}_0 = \begin{bmatrix}
m I_3 & 0 \\
0 & \vect{J}
\end{bmatrix}\dot{\bm{y}}_0 -\vect{F} \vect{u}_0\label{eq:measured_f}.
\end{equation}
Inserting \eqref{eq:measured_f} into \eqref{eq:NDIcontrol_law} the fully actuated INDI control law is obtained by
\begin{equation}
\vect{u} = \vect{F}^{-1}\begin{bmatrix}
m I_3 & 0 \\
0 & \vect{J}
\end{bmatrix}\!\left(
- \dot{\bm{y}}_0  +
\vect{\nu}_{d}
\right) + \bm{u}_0,
\label{eq:INDIcontrol_law}
\end{equation}

where $\dot{\bm{y}}_0 $ denotes measured translational and rotational acceleration. The translational acceleration measurements are directly available through the onboard accelerometer; however, the common UAV platforms lack a rotational acceleration sensor. Instead, the rotational velocity is measured by the onboard gyroscope, then passed through a second-order low-pass filter and differentiated. The same second-order filter is applied to the actuator and translational acceleration feedback to achieve synchronization \cite{STEINERT2025103591}.

$\bm{\nu}_{d}$ denotes the desired pseudo control. It is generated by an outer-loop controller based on the approach of Rupprecht et al. \cite{timOuterref}. In the outer loop, a reference model is used to generate physically feasible commands, and an error controller ensures they are tracked. 

$\bm{u}_0$ denotes the squared propeller rotation speeds, obtained from direct sensor measurements.

\begin{figure*}[t]
    \centering
    \footnotesize
    \vspace{-2mm}
    \scalebox{0.82}{%
    \begin{tikzpicture}[
        x=0.95cm,y=0.95cm,
        block/.style={draw, rounded corners=2pt, align=center, minimum width=2.35cm, minimum height=0.85cm, fill=gray!8},
        io/.style={draw, rounded corners=2pt, align=center, minimum width=2.0cm, minimum height=0.7cm, fill=white},
        ctrl/.style={draw, rounded corners=2pt, align=center, minimum width=2.35cm, minimum height=0.75cm, thick, font=\bfseries},
        geostyle/.style={ctrl, draw=blue!65!black, fill=blue!10},
        indistyle/.style={ctrl, draw=orange!85!black, fill=orange!12},
        line/.style={->, thick},
        sig/.style={midway, fill=white, inner sep=1.2pt}
    ]
        \draw[rounded corners=4pt, draw=blue!65!black, fill=blue!4, thick] (1.7,0.85) rectangle (8.90,2.05);
        \draw[rounded corners=4pt, draw=orange!85!black, fill=orange!6, thick] (1.7,-2.1) rectangle (8.90,-0.75);

        \node[font=\bfseries\scriptsize, text=red!80!black] at (5.3,2.30) {either};
        \node[font=\bfseries\scriptsize, text=red!80!black] at (5.3,-0.50) {or};
        \node[font=\bfseries\scriptsize, text=blue!60!black] at (7.6,2.30) {GEO NDI};
        \node[font=\bfseries\scriptsize, text=orange!80!black] at (7.9,-.5) {INDI};

        \node[io] (ref) at (-1,0) {Reference\\commands};
        \node[block] (outer) at (2.2,0) {Shared outer loop\\error dynamics\\and pseudo control};

        \node[geostyle] (model) at (3.2,1.45) {Model\\information};
        \node[indistyle] (sensor) at (3.2,-1.45) {Sensor\\feedback};

        \node[geostyle] (geo) at (7.4,1.45) {Model-based\\inversion (Eq.\ref{eq:NDIcontrol_law})};
        \node[indistyle] (indi) at (7.4,-1.45) {Sensor-based\\inversion (Eq.\ref{eq:INDIcontrol_law})};

        \node[block, minimum width=1.85cm, minimum height=0.62cm] (alloc) at (10.9,0) {Control\\allocation};
        \node[block, minimum width=1.85cm, minimum height=0.62cm] (plant) at (13.9,0) {Platform};
        \node[io] (state) at (13.9,-2.45) {State\\estimation};

        \draw[line] (ref) -- node[sig, above, yshift=1.5pt] {$\vect{p}_d,\vect{q}_d$} (outer);
        \path (outer.east) ++(3.9,0) coordinate (outerbranch);
        \draw[line] (outer.east) -- node[sig, above] {$\vect{\nu}_d$} (outerbranch) -| (geo.south);
        \draw[line] (outer.east) --  (outerbranch) -| (indi.north);
        \draw[line] (model.east) -- node[sig, above, fill=blue!4] {$\vect{F},\vect{f},m,\vect{J}$} (geo.west);
        \draw[line] (sensor.east) -- node[sig, below, fill=orange!6, xshift=2pt] {$\dot{\vect{y}}_0,u_0$} (indi.west);
        \draw[line] (geo.east) -- ++(0.55,0) |- (alloc.west);
        \draw[line] (indi.east) -- ++(0.55,0) |- node[pos=.55, above, xshift=2pt] {$u$} (alloc.west);
        \draw[line] (alloc) -- node[sig, above] {RPM} (plant);
        \draw[line] (plant) -- (state);
        \draw[line] (state.west) -- ++(-11.39,0)
  node[pos=.5, below] {$\vect{p},\vect{q},\vect{v},\vect{\omega}$}
  -| ([xshift=-7mm]outer.south);
    \end{tikzpicture}}
    \caption{Overall control structure used for both controllers. The shared outer loop generates pseudo-control commands, while the colored GEO and INDI branches highlight the two alternative inversion strategies. Signal symbols are shown on the arrows to emphasize how the two branches use different information sources.}
    \vspace{-3mm}
    \label{fig:control_flowchart}
\end{figure*}
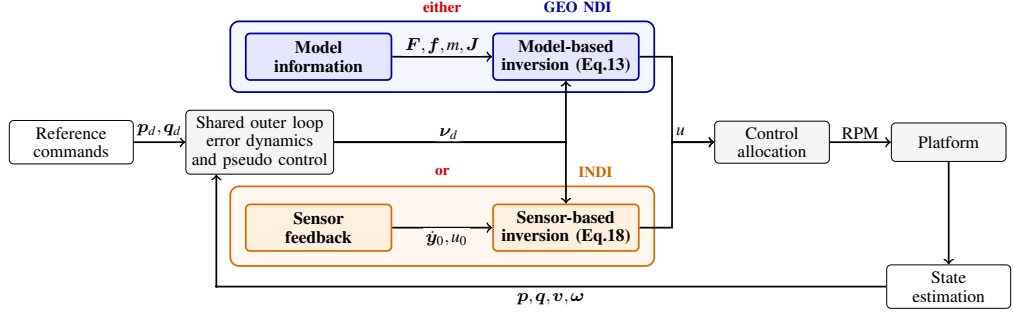

\section{Platform}

\begin{figure}
    \centering
    \includegraphics[width=0.5\linewidth]{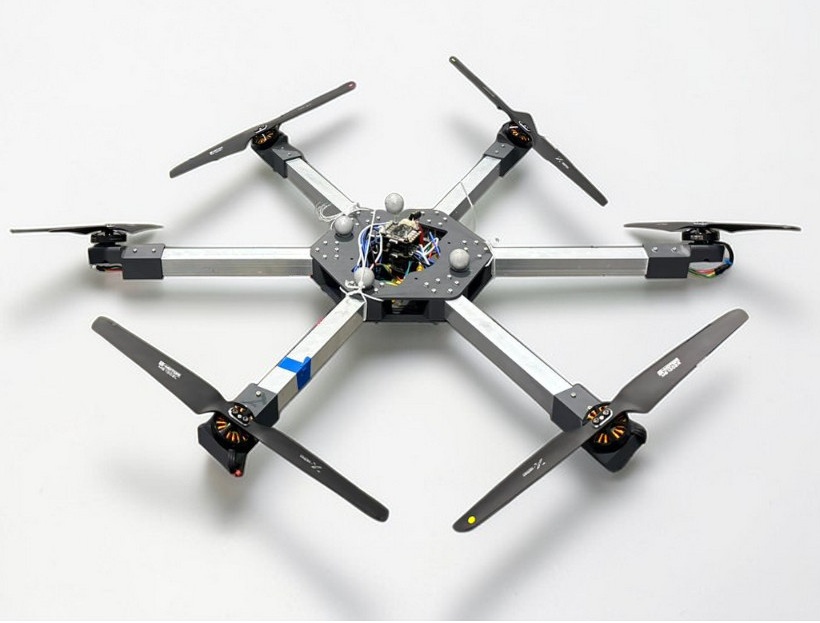}
    \caption{Photograph of the fixed-tilt fully actuated platform used in the experimental campaign.}
    \label{fig:cadl}
\end{figure}

The experiments are carried out on a fixed-tilt fully actuated hexarotor. This platform is chosen because of its simple mechanical design and ease of manufacturing. The rotors are mounted at fixed tilt angles relative to their arms radial axis as shown in Fig. \ref{fig:cadl}. This removes the need for additional tilting mechanisms and keeps the system lightweight and robust.

The total mass of the platform, including the battery, is 2.95 kg. The distance between opposite motor shafts is 750 mm. The vehicle uses 13×4.5 inch propellers, which provide a good balance between efficiency and control authority. The rotor speeds are limited between $8-100~Hz$ to prevent heating and ensure actuator longevity. The propulsion system achieves a maximum thrust-to-weight ratio of 2.8, allowing the platform to generate large forces and moments when required.

All rotors are mounted with a fixed tilt angle of 30°, enabling the generation of lateral forces without changing the vehicle attitude. 

The system uses a dual-computer architecture. A Kakute H7 flight controller handles low-level tasks such as sensor measurements and motor control. High-level control algorithms run on an x86-based companion computer.

The sensor suite consists of the onboard IMU of the Kakute H7 flight controller, which provides specific force and angular velocity measurements at \SI{500}{\hertz}, and a Vicon motion capture system, which supplies position and attitude measurements at \SI{250}{\hertz} with submillimeter accuracy. The sensor measurements are fused using an Extended Kalman Filter to enable full-state feedback, including velocity estimation.

\begin{figure*}[t]
    \centering
    \includegraphics[width=1\linewidth]{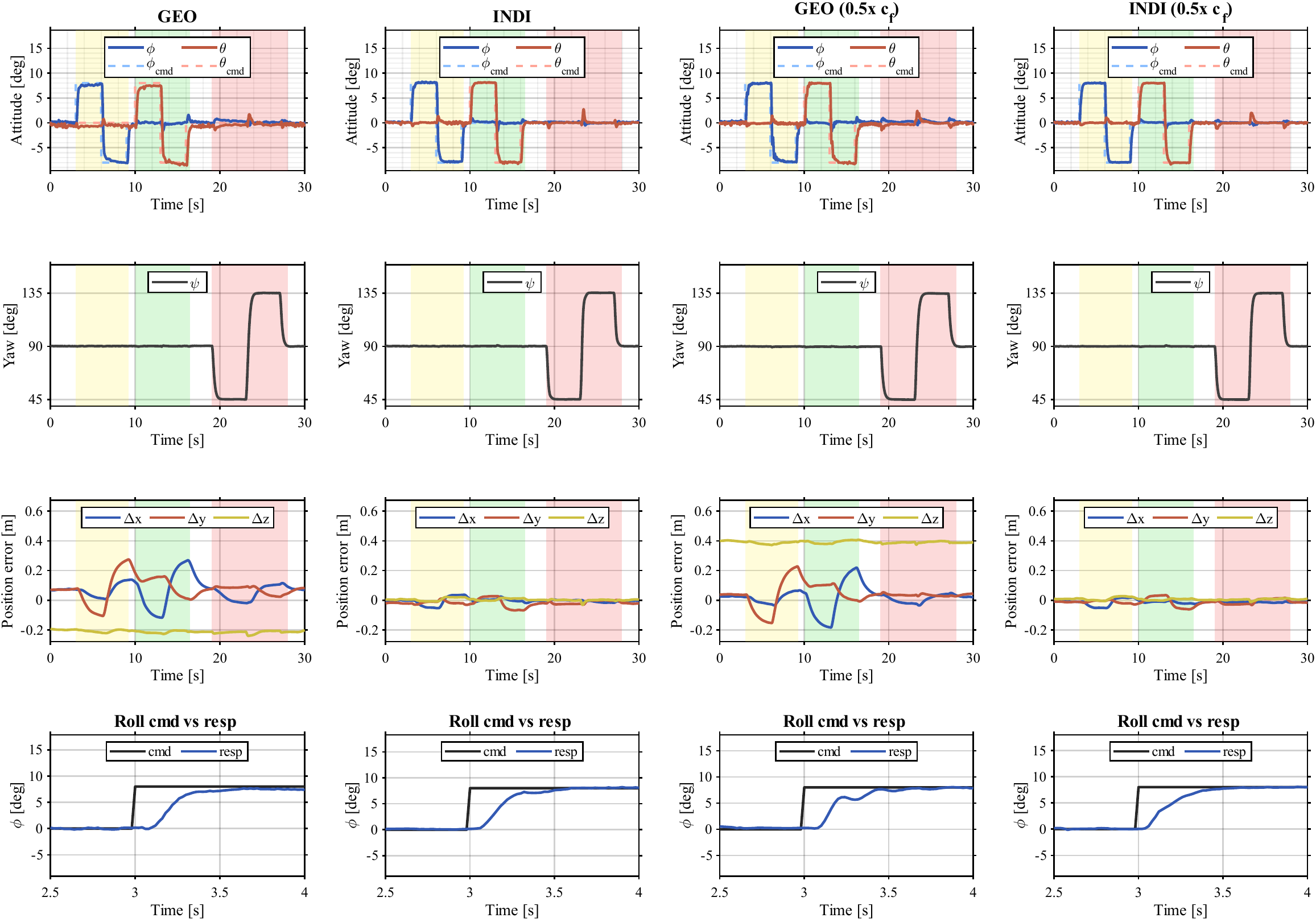}
    \caption{Attitude and position tracking results for the geometric NDI (GEO) and INDI controllers under nominal parameters and with a 50\% reduction of the rotor force coefficient $c_f$. Columns (left to right): GEO (nominal), INDI (nominal), GEO with param. mismatch, and INDI with with param. mismatch. 
Rows (top to bottom): roll and pitch tracking, yaw tracking, position tracking error, and a zoomed-in roll response for a representative step command of $8^\circ$. Highlighted regions indicate the active command axis (yellow: roll, green: pitch, red: yaw).}
    \label{fig:A}
\end{figure*}

\section{Experiments}
To enable a direct and fair comparison between the model-based geometric NDI and incremental NDI controllers, both methods were implemented with identical translational (position and velocity) and rotational (attitude and angular rate) control gains. This ensures that both controllers generate commanded translational and rotational accelerations from the same error dynamics, such that any observed performance differences originate from the inversion strategy rather than from discrepancies in gain tuning. In addition, both controllers were executed at \SI{500}{\hertz}, which corresponds to the communication bandwidth limit of the experimental platform. This update rate is higher than that typically reported in industrial and research multirotor flight control systems, ensuring that both approaches operate under identical high-bandwidth timing conditions. 
In total, five experiments were conducted. Experiments~1--3 evaluate parameter uncertainty, hover disturbance rejection, and waypoint tracking under gusts. Experiments ~4 and~5 provide complementary robustness studies on reduced control frequency and injected sensor degradation. 
An accompanying experiment video is provided in the first-page footnote.
\subsection{Experiment 1: Parameter Uncertainty}

For this experiment, two tests were conducted for each controller. In the first test, the model parameters were kept at their nominal values to evaluate baseline tracking performance. In the second test, the rotor force coefficient $c_f$ in \eqref{eq:rotor_force} used by both controllers was reduced to 50 $\%$ of its nominal value in order to introduce a controller parameter mismatch. The platform was commanded to hover at a fixed position and attitude setpoint, after which sequential attitude step commands were applied.
\begin{figure*}
    \centering
    \includegraphics[width=0.9\linewidth]{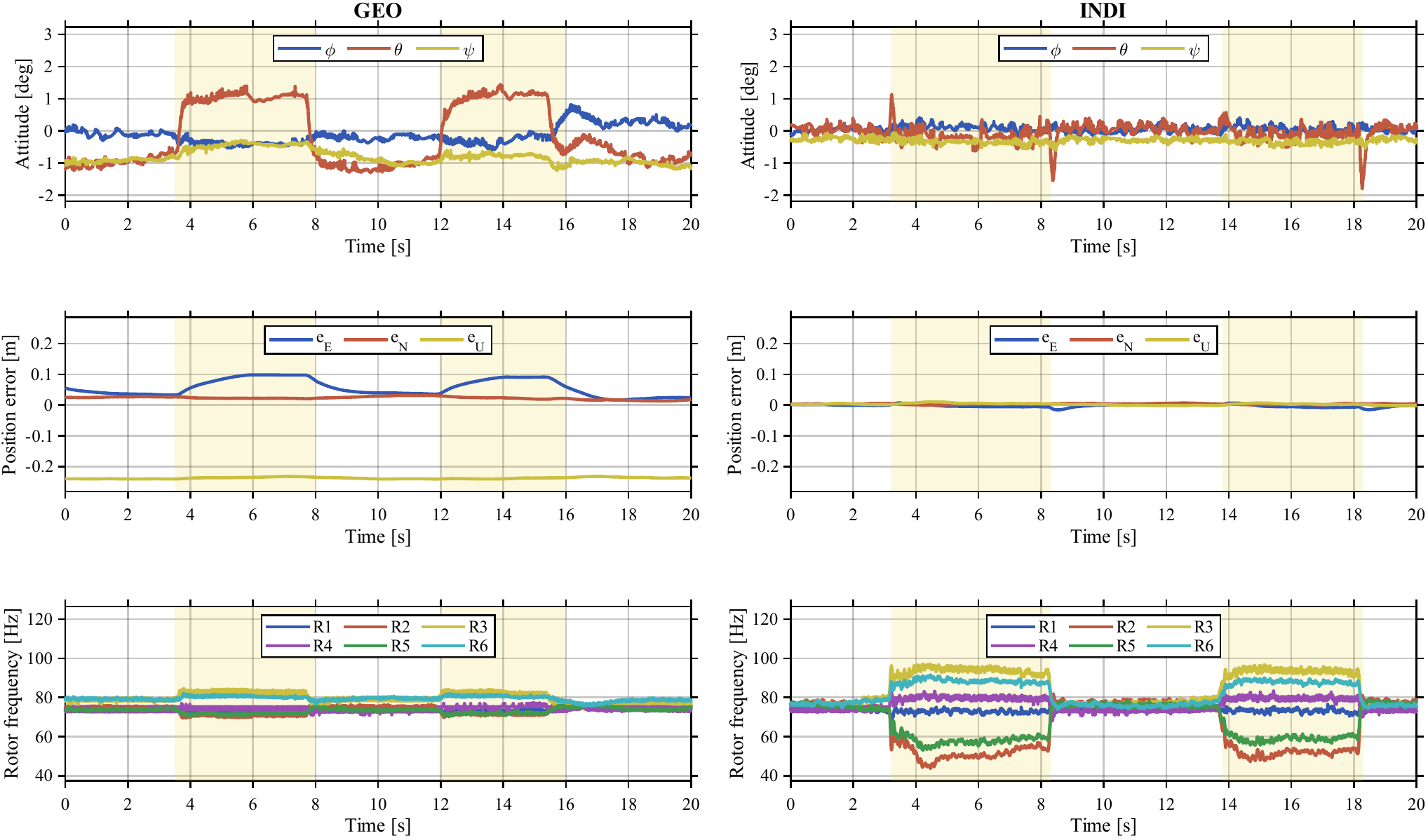}
    \caption{Attitude and position tracking results for the geometric NDI (GEO) and INDI controllers under constant force and torque test. Columns (left to right): GEO , INDI. Rows (top to bottom): Attitude tracking, position tracking error, and rotor rotational speeds. Highlighted regions indicate the active load.}
    \label{fig:GEO_INDI_Load_Test}
\end{figure*}
Fig. \ref{fig:A} shows the attitude tracking performance of the controllers under nominal conditions and parameter mismatch. The maneuver consists of sequential attitude step commands of $\pm8^\circ$ in roll and pitch and $\pm 45 ^ \circ$ in yaw. The highlighted regions indicate the active command axis: yellow corresponds to roll, green to pitch, and red to yaw. The fourth row provides close-up figures that highlight the response.\\

Both nominal controllers maintain stable flight and successfully track the commanded maneuvers in all cases. However, INDI consistently achieves more accurate attitude tracking than the geometric controller with NDI. Under nominal conditions, INDI closely follows the commanded steps in all three axes. However, the geometric control approach exhibits small but noticeable steady-state attitude errors, as well as higher position tracking errors during maneuvers. The roll close-up figure clearly demonstrates this behaviour, showing a slight bias between command and response, with similar offsets present in the other attitude axes as well. 

The difference between the controllers becomes more pronounced when the rotor force coefficient is reduced to $0.5\ c_f$. The geometric NDI responses show increased tracking error and slower convergence across all maneuver phases in attitude, indicating sensitivity to the force-coefficient mismatch in the model-based inversion. In contrast, the INDI responses remain close to the nominal response, with only minor transient deviations and no noticeable steady-state offset. The roll close-up confirms that INDI maintains similar rise time and accuracy despite the parameter change, whereas the geometric controller's response becomes slower and less accurate. The rise time performance of INDI is qualitatively shown in Table \ref{tab:TestA}.  

Overall, the results demonstrate that although both controllers stabilize the vehicle under significant uncertainty in the rotor force coefficient, INDI achieves consistently better attitude tracking accuracy and robustness than geometric NDI. This behavior is expected, since INDI updates the control input based on measured response and therefore is less sensitive to errors in the assumed force coefficient. A qualitative analysis of the performance difference of the controllers is depicted in  Table \ref{tab:TestA}. The repeated-test statistics over 10 runs, summarized in Table \ref{tab:all_cases_norm_summary}, confirm the same trend. In this table, the longitudinal attitude error, defined using roll and pitch, and the position error are reported.

\begin{table}
    \centering
    \caption{Experiment 1: Metrics for Parameter Uncertainty.}
    \vspace{-3mm}
    \label{tab:TestA}
    \setlength{\tabcolsep}{3pt}
    \renewcommand{\arraystretch}{1.15}
    \scriptsize
    \resizebox{\columnwidth}{!}{%
    \begin{tabular}{|>{\columncolor{gray!15}}l|c|c|c|c|}
        \hline
        \rowcolor{gray!15}
        \textbf{Metric} & \textbf{GEO} & \textbf{INDI} & \textbf{GEO (0.5x $c_f$)} & \textbf{INDI (0.5x $c_f$)} \\
        \hline
        $|\Delta x|$ mean [m] & 0.08 & 0.01 & 0.05 & 0.02 \\
        \hline
        $|\Delta y|$ mean [m] & 0.09 & 0.02 & 0.06 & 0.02 \\
        \hline
        $|\Delta z|$ mean [m] & 0.21 & 0.01 & 0.39 & 0.01 \\
        \hline
        $|\Delta x|$ peak [m] & 0.27 & 0.05 & 0.22 & 0.05 \\
        \hline
        $|\Delta y|$ peak [m] & 0.27 & 0.07 & 0.23 & 0.06 \\
        \hline
        $|\Delta z|$ peak [m] & 0.24 & 0.03 & 0.41 & 0.03 \\
        \hline
        $|e_{\phi}|$ mean [deg] & 0.49 & 0.36 & 0.45 & 0.30 \\
        \hline
        $|e_{\theta}|$ mean [deg] & 0.77 & 0.34 & 0.47 & 0.33 \\
        \hline
        $|e_{\psi}|$ mean [deg] & 2.37 & 1.81 & 2.31 & 1.74 \\
        \hline
        roll rise time [s] & 0.28 & 0.24 & 0.35 & 0.27 \\
        \hline
    \end{tabular}%
    }
\end{table}

\begin{table}
    \centering
    \caption{Aggregate norm error statistics across all repeated-test cases.}
    \vspace{-3mm}
    \label{tab:all_cases_norm_summary}
    \renewcommand{\arraystretch}{1.15}
    \resizebox{\columnwidth}{!}{%
    \begin{tabular}{|>{\columncolor{gray!15}}l|c|c|}
        \hline
        \rowcolor{gray!15}
        \textbf{Case} & \textbf{Longitudinal attitude error [deg]} & \textbf{Position error [m]} \\
        \hline
        INDI & 0.58 $\pm$ 1.80 & 0.0074 $\pm$ 0.0050 \\
        \hline
        GEO & 1.17 $\pm$ 1.73 & 0.2520 $\pm$ 0.0290 \\
        \hline
        INDI (0.5x $c_f$) & 0.54 $\pm$ 1.76 & 0.0111 $\pm$ 0.0053 \\
        \hline
        GEO (0.5x $c_f$) & 0.75 $\pm$ 1.63 & 0.4033 $\pm$ 0.0187 \\
        \hline
    \end{tabular}
    }
\end{table}
\subsection{Experiment 2: Hover Disturbance Rejection}

To conduct this experiment, the platform was commanded to hover at a fixed position and horizontal attitude setpoint. While this condition was maintained, an external disturbance was applied. The disturbance was introduced using a hook mounted beneath the rear arm of the platform ($z$-dir distance of $0.13$m), to which a pulley system with a $ 500$ $g$ mass was attached via a string. The string was initially slack to ensure that the platform was not affected during the undisturbed hover state. The mass was then released, applying a lateral $4.905~N$ load and a $0.638Nm$ pitch moment due to the vertical offset. The disturbance was subsequently removed by quickly lifting the mass, causing the string to become slack again.

In Fig. \ref{fig:GEO_INDI_Load_Test}, the test results for the hover disturbance rejection for both controllers are given.

A clear difference in disturbance rejection is evident between the two controllers. For INDI, the disturbance causes only a brief attitude deviation of about $\pm1 ^ \circ$ at the moments when the load is applied and removed. After this initial impact, the controller quickly damps the motion and returns the vehicle to horizontal hover. The position error remains small. The peak and mean position and attitude error can be seen in Table \ref{tab:TestB}. The rotor speeds quickly rise to reduce the effects of the external disturbance.

The geometric NDI controller exhibits reduced accuracy under the same disturbance. The pitch angle shows relatively large deviations that persist during load application. The lateral position errors change significantly, indicating that the controller is struggling to maintain position during the disturbance. The platform also takes longer to return to the horizontal hover state after the load is removed, with visible residual motion. The rotor speeds change less aggressively compared to the INDI controller.

Overall, this experiment demonstrates the performance differences between the two controllers. INDI limits the disturbance effect to a short transient at impact and rapidly suppresses it thereafter. In contrast, the geometric controller exhibits sustained attitude and position deviations during loading. These results indicate the stronger disturbance rejection capability of INDI, which directly compensates for external forces using measured motion, whereas geometric NDI is more sensitive to unmodeled disturbances.

\begin{table}
    \centering
    \caption{Experiment 2: Metrics for hover disturbance rejection.}
    \vspace{-2mm}
    \label{tab:TestB}
    \renewcommand{\arraystretch}{1.15}
    \begin{tabular}{|>{\columncolor{gray!15}}l|c|c|}
        \hline
        \rowcolor{gray!15}
        \textbf{Metric} & \textbf{GEO} & \textbf{INDI} \\
        \hline
        $|\Delta x|$ mean [m] & 0.055 & 0.004 \\
        \hline
        $|\Delta y|$ mean [m] & 0.023 & 0.003 \\
        \hline
        $|\Delta z|$ mean [m] & 0.238 & 0.003 \\
        \hline
        $|\Delta x|$ peak [m] & 0.098 & 0.016 \\
        \hline
        $|\Delta y|$ peak [m] & 0.031 & 0.007 \\
        \hline
        $|\Delta z|$ peak [m] & 0.241 & 0.011 \\
        \hline
        $|e_{\phi}|$ mean [deg] & 0.26 & 0.09 \\
        \hline
        $|e_{\theta}|$ mean [deg] & 0.93 & 0.18 \\
        \hline
        $|e_{\psi}|$ mean [deg] & 0.22 & 0.70 \\
        \hline
        $|e_{\phi}|$ peak [deg] & 0.82 & 0.42 \\
        \hline
        $|e_{\theta}|$ peak [deg] & 1.44 & 1.80 \\
        \hline
        $|e_{\psi}|$ peak [deg] & 0.71 & 0.95 \\
        \hline
    \end{tabular}
    \vspace{-5mm}
\end{table}

\subsection{Experiment 3: Waypoint Tracking Under Gust}
In this experiment, the platform was commanded to follow a set of waypoints, during which a wind gust disturbance was applied using a fan and an additional light plate attached beneath the platform to exacerbate the disturbance effects.

\begin{figure}
    \centering
    \includegraphics[width=0.9\linewidth]{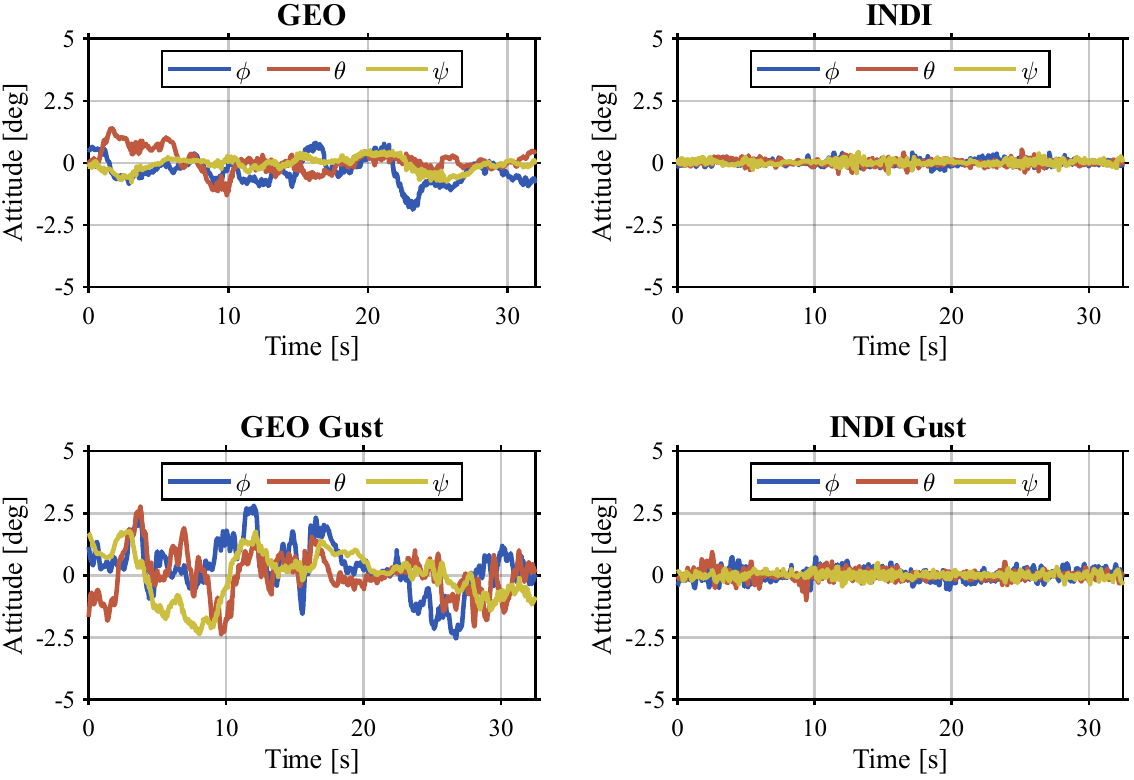}
    \caption{Attitude tracking results for the geometric NDI (GEO) and INDI controllers under waypoint tracking test with and without gusts. Columns (left to right): no gusts and gust conditions. Rows (top to bottom): GEO, INDI.}
    \label{fig:Waypoint}
    \vspace{-6mm}
\end{figure}

The experiments show that both controllers achieved similar waypoint-tracking performance, with no significant differences observed in the position error signals. For this reason, position results are not reported. However, clear differences were seen in the attitude response required to counteract the disturbances. Accordingly, Fig. \ref{fig:Waypoint} focuses on the attitude behavior during the experiment. Under gust conditions, the geometric controller exhibits larger attitude deviations and more oscillatory behavior, whereas the INDI controller maintains more robust attitude regulation. These results indicate that although both controllers achieve comparable waypoint tracking, INDI provides more effective disturbance rejection at the attitude level. The qualitative performance analysis of the Fig. \ref{fig:Waypoint} can be seen in Table \ref{tab:waypoint}.

\begin{table}
    \centering
    \caption{Experiment 3: Attitude tracking error statistics under nominal and gust disturbances.}
    \label{tab:waypoint}
    \renewcommand{\arraystretch}{1.15}
    \begin{tabular}{|>{\columncolor{gray!15}}l|c|c|c|c|}
        \hline
        \rowcolor{gray!15}
        \textbf{Metric} & \textbf{INDI} & \textbf{INDI (gust)} & \textbf{GEO} & \textbf{GEO (gust)} \\
        \hline
        $|e_{\phi}|$ mean [deg]   & 0.08 & 0.18 & 0.49 & 0.80 \\
        \hline
        $|e_{\theta}|$ mean [deg] & 0.09 & 0.13 & 0.34 & 0.68 \\
        \hline
        $|e_{\psi}|$ mean [deg]   & 0.07 & 0.09 & 0.22 & 0.80 \\
        \hline
        $|e_{\phi}|$ peak [deg]   & 0.42 & 0.75 & 1.89 & 2.81 \\
        \hline
        $|e_{\theta}|$ peak [deg] & 0.55 & 1.01 & 1.40 & 2.77 \\
        \hline
        $|e_{\psi}|$ peak [deg]   & 0.44 & 0.50 & 0.78 & 2.36 \\
        \hline
    \end{tabular}
    \vspace{-3mm}
\end{table}
\subsection{Experiment 4: Different Control Frequencies}
In this experiment, the hovering task was repeated while reducing the controller execution frequency from the nominal 500~Hz to 250~Hz, 125~Hz, 62.5~Hz, and 50~Hz. The purpose was to investigate how strongly both inversion strategies depend on update rate when applied to a fully actuated platform. Since INDI relies directly on high-rate sensor feedback and incremental control updates, a stronger sensitivity to reduced sampling frequency is expected than for the geometric controller.

\begin{figure}[!t]
    \centering
    \includegraphics[width=0.9\linewidth]{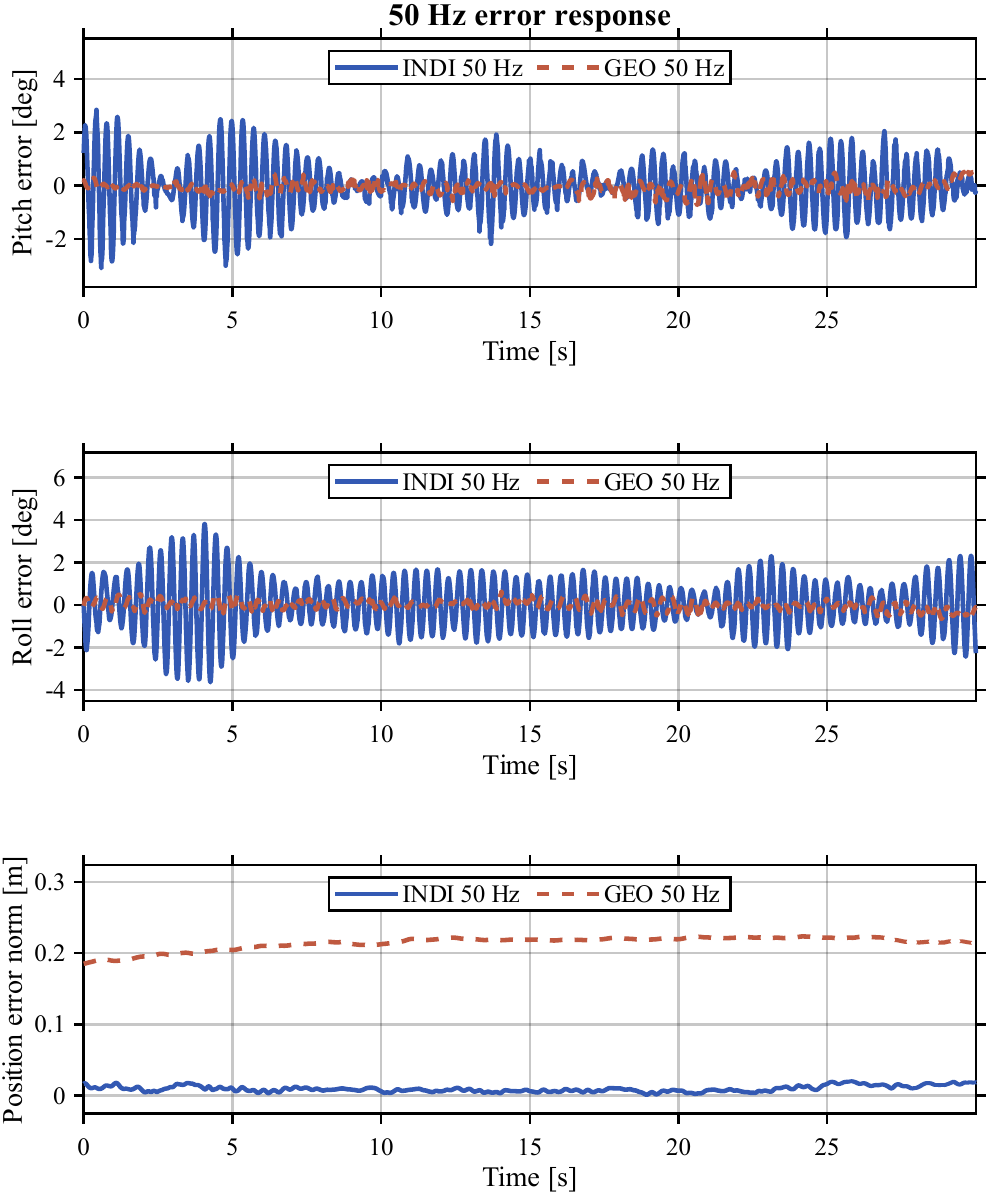}
    \caption{Attitude and position error results for the GEO and INDI controllers under the 50 Hz controller frequency test. Rows from top to bottom show pitch error, roll error, and position error norm; each subplot compares GEO and INDI over the selected test interval.}
    \label{fig:freq_test}
\end{figure}

The results confirm the expectation. With a decreasing control frequency, the attitude tracking performance of INDI degrades significantly. This trend is especially visible in Fig. \ref{fig:freq_test}, which plots the 50~Hz case. Under this condition, the INDI-controlled platform can no longer maintain the same attitude quality observed at higher rates, and noticeably higher roll and pitch errors appear. In contrast, the geometric controller remains stable throughout the frequency sweep and preserves a comparatively smoother attitude response at 50~Hz. The aggregate statistics in Table~\ref{tab:frequency_shift_norm_summary} reflect the same trend: the INDI longitudinal attitude error increases from 0.19 $\pm$ 0.11 deg at 500~Hz to 1.34 $\pm$ 0.68 deg at 50~Hz, whereas the GEO longitudinal error remains bounded and comparatively small at 0.26 $\pm$ 0.15 deg in the 50~Hz case. 

Despite this attitude degradation, INDI still regulates position better than the geometric controller across all tested frequencies. As shown in Fig.~\ref{fig:freq_test} and Table~\ref{tab:frequency_shift_norm_summary}, at 50~Hz INDI achieves a position error of 0.0095 $\pm$ 0.0041m, whereas GEO reaches 0.2144 $\pm$ 0.0092m. Thus, reduced control frequency affects the attitude channel of INDI more strongly, while its translational performance remains clearly superior. 
\begin{table}[!t]
\centering
\caption{Controller frequency norm error statistics for INDI and GEO.}
\vspace{-2mm}
\label{tab:frequency_shift_norm_summary}
\renewcommand{\arraystretch}{1.05}
\setlength{\tabcolsep}{3pt}
\resizebox{\columnwidth}{!}{
\begin{tabular}{|>{\columncolor{gray!15}}l|c|c|c|c|}
\hline
\rowcolor{gray!15}
\textbf{Freq.} & \textbf{INDI lon. [deg]} & \textbf{INDI pos. [m]} & \textbf{GEO lon. [deg]} & \textbf{GEO pos. [m]} \\
\hline
50 Hz   & 1.34 $\pm$ 0.68 & 0.0095 $\pm$ 0.0041 & 0.26 $\pm$ 0.15 & 0.2144 $\pm$ 0.0092 \\
\hline
62.5 Hz & 0.24 $\pm$ 0.12 & 0.0119 $\pm$ 0.0042 & 0.43 $\pm$ 0.15 & 0.2693 $\pm$ 0.0144 \\
\hline
125 Hz  & 0.15 $\pm$ 0.10 & 0.0069 $\pm$ 0.0019 & 0.49 $\pm$ 0.22 & 0.2423 $\pm$ 0.0095 \\
\hline
250 Hz  & 0.11 $\pm$ 0.09 & 0.0075 $\pm$ 0.0024 & 0.81 $\pm$ 0.23 & 0.2338 $\pm$ 0.0090 \\
\hline
500 Hz  & 0.19 $\pm$ 0.11 & 0.0264 $\pm$ 0.0050 & 0.76 $\pm$ 0.21 & 0.2158 $\pm$ 0.0111 \\
\hline
\end{tabular}
}
\vspace{-3mm}
\end{table}

\subsection{Experiment 5: Hover Sensor Degradation}
For this experiment, additional noise was injected into the controllers, and the platform was again commanded to hover at a fixed position and horizontal attitude setpoint. The aim was to compare the performance of the controllers under unexpected sensor degradation as an external disturbance. For the INDI controller, noise was injected into both the translational acceleration and angular velocity feedback, while for NDI, noise was injected only into the angular velocity channel, as acceleration measurements are not directly used in the control law. The noise was injected directly into the controller feedback to prevent state estimation performance from affecting the results.

Table \ref{tab:sensor_degradation_norm_summary} presents the sensor-degradation error statistics for both controllers in hover. The leftmost column indicates the injected noise covariance level, selected based on the onboard sensor specifications. The angular columns indicate deviations from the horizontal plane (excluding yaw), while the position columns indicate deviations from the position setpoint. All results are reported in terms of the norm and standard deviation.

The results show that the INDI controller maintains superior performance even at high noise levels, with lower norms across all noise levels for both angular and position errors. The standard deviations of the angular errors are generally slightly higher for the geometric controller, while the position standard deviations are higher for the INDI controller. This behavior is expected, as the geometric controller does not use direct accelerometer measurements.

\begin{table}[!t]
    \centering
    \vspace{-2mm}
    \caption{Sensor-degradation norm error statistics for INDI and GEO.}
    \label{tab:sensor_degradation_norm_summary}
    \renewcommand{\arraystretch}{1.1}
    \setlength{\tabcolsep}{3pt} 
    \scriptsize
    \begin{tabular}{|>{\columncolor{gray!15}}l|c|c|c|c|}
        \hline
        \rowcolor{gray!15}
        Noise & INDI Ang. & INDI Pos. & GEO Ang. & GEO Pos. \\
        \hline
        $0\times\sigma^2$  & 0.16 $\pm$ 0.10 & 0.0197 $\pm$ 0.0057 & 0.68 $\pm$ 0.16 & 0.2157 $\pm$ 0.0098 \\
        \hline
        $1\times\sigma^2$  & 0.26 $\pm$ 0.24 & 0.0139 $\pm$ 0.0022 & 0.53 $\pm$ 0.26 & 0.2297 $\pm$ 0.0025 \\
        \hline
        $3\times\sigma^2$  & 0.35 $\pm$ 0.22 & 0.0166 $\pm$ 0.0059 & 0.63 $\pm$ 0.35 & 0.2183 $\pm$ 0.0034 \\
        \hline
        $7\times\sigma^2$  & 0.46 $\pm$ 0.31 & 0.0288 $\pm$ 0.0055 & 0.69 $\pm$ 0.46 & 0.2164 $\pm$ 0.0029 \\
        \hline
        $15\times\sigma^2$ & 0.63 $\pm$ 0.46 & 0.0408 $\pm$ 0.0055 & 0.83 $\pm$ 0.56 & 0.2244 $\pm$ 0.0046 \\
        \hline
        $31\times\sigma^2$ & 0.77 $\pm$ 0.52 & 0.0441 $\pm$ 0.0108 & 0.98 $\pm$ 0.65 & 0.2371 $\pm$ 0.0032 \\
        \hline
    \end{tabular}
        \vspace{-2mm}

\end{table}

\subsection{Discussion}
The superior performance of INDI compared to NDI observed in the experiments can be explained by the theoretical analysis presented by Wang et al. \cite{Wang19}. In Experiment 1, the rotor force coefficient $c_f$ was altered, which can be treated as a model discrepancy in the control effectiveness matrix $\vect{F}$ shown in \eqref{eq:model}. As explained in their work, such model differences affect the closed-loop response as perturbations. Under these conditions, INDI exhibits a much smaller perturbation norm than NDI. Moreover, for INDI, this norm can be further tightened by increasing the sampling rate of the controller.

In Experiments 2, 3, and 5, the platform is affected by external disturbances. As discussed in their analysis, for INDI, the steady component of the disturbance is already included in the measurements and compensated by the controller, while only variations in the disturbance affect the error dynamics. Moreover, these effects can be further reduced by increasing the sampling rate. This is not the case for NDI, for which any unmodeled disturbance affects the error dynamics directly and is only compensated by the error controller. Experiment~4 further supports this interpretation. At lower sampling rates, INDI exhibits exacerbated degradation in attitude performance compared to the geometric controller, while still maintaining superior translational performance.
\vspace{-2mm}
\section{Conclusion}
This paper presented an experimental comparison between model-based geometric nonlinear dynamic inversion (NDI) and sensor-based incremental nonlinear dynamic inversion (INDI) for a fixed-tilt fully actuated hexarotor. To ensure a fair comparison, both controllers used the same translational and rotational error dynamics and were executed at 500~Hz; thus, the observed differences primarily reflect the inversion mechanism rather than tuning.

Overall, for fully actuated multirotors where position and attitude can be controlled independently, the five experiments show that INDI provides clear advantages when model parameters or sensor quality are uncertain and when disturbance rejection in translation is critical. At the same time, these advantages are not universal: geometric NDI is less sensitive to reduced control frequency in the attitude channel. Future work will focus on performance analysis of INDI on underactuated and fully actuated platforms, as well as improving the robustness of both controllers.
\vspace{-2mm}
\section*{Acknowledgment}
This work was funded by the Deutsche Forschungsgemeinschaft (DFG, German Research Foundation) under project BE 4791/7.
\vspace{-4mm}

\addtolength{\textheight}{-0cm}   




\bibliographystyle{IEEEtran.bst}
\bibliography{IEEEexample}

\end{document}